\pdfoutput=1
\documentclass{article}

\usepackage{arxiv}

\usepackage[utf8]{inputenc} 
\usepackage[T1]{fontenc}    
\usepackage{hyperref}       
\usepackage{url}            
\usepackage{booktabs}       
\usepackage{amsfonts}       
\usepackage{nicefrac}       
\usepackage{microtype}      
\usepackage{lipsum}
\usepackage{booktabs} 
\usepackage{multirow}
\usepackage[flushleft]{threeparttable}
\usepackage{adjustbox}
\usepackage{graphicx}
\usepackage{subcaption}
\usepackage{verbatim}
\usepackage{authblk}

\title{Reasoning over RDF Knowledge Bases using Deep Learning}

\author[1]{Monireh Ebrahimi}
\author[1]{Md Kamruzzaman Sarker}
\author[1,2]{Federico Bianchi}
\author[1]{Ning Xie}
\author[1]{Derek Doran}
\author[1]{Pascal Hitzler}
\affil[1]{Department of Computer Science \& Engineering, Wright State University}
\affil[2]{University of Milan - Bicocca}
\affil[1]{\textit {\{ebrahimi.2, sarker.3, xie.25, derek.doran, pascal.hitzler\}@wright.edu}}
\affil[2]{\textit {federico.bianchi@disco.unimib.it}}

\begin{document}
\maketitle

\begin{abstract}
Semantic Web knowledge representation standards, and in particular RDF and OWL, often come endowed with a formal semantics which is considered to be of fundamental importance for the field. Reasoning, i.e., the drawing of logical inferences from knowledge expressed in such standards, is traditionally based on logical deductive methods and algorithms which can be proven to be sound and complete and terminating, i.e. correct in a very strong sense. 

For various reasons, though, in particular the scalability issues arising from the ever increasing amounts of Semantic Web data available and the inability of deductive algorithms to deal with noise in the data, it has been argued that alternative means of reasoning should be investigated which bear high promise for high scalability and better robustness. From this perspective, deductive algorithms can be considered the gold standard regarding correctness against which alternative methods need to be tested. 

In this paper, we show that it is possible to train a Deep Learning system on RDF knowledge graphs, such that it is able to perform reasoning over new RDF knowledge graphs, with high precision and recall compared to the deductive gold standard.
\end{abstract}

\keywords{deep learning \and reasoning \and knowledge base}

\section{Introduction}
Automated reasoning, attempting to conduct logical reasoning algorithmically, has been a long-standing focus of general artificial intelligence with wide application in knowledge base completion, natural language understanding, question answering, agent planning and etc. In particular, with the recent advent of web-scale knowledge graphs, such as Freebase\cite{bollacker2008freebase}, DBpedia \cite{lehmann2015dbpedia},  and Google's Knowledge Vault \cite{dong2014knowledge} and due to their high incompleteness \cite{min2013distant} there has been rising interest in solutions to the knowledge base completion task.

For many years, reasoning has been tackled as the task of building systems capable of inferring new crisp symbolic logical rules \cite{mccarthy1960programs,nilsson1991logic}. However, those traditional methods are too brittle to be applied to noisy automatically created knowledge bases. With the recent revival of interest in artificial neural networks, neural link prediction models have been applied vastly for the completion of knowledge graphs. These methods \cite{nickel2012factorizing,riedel2013relation,socher2013reasoning,chang2014typed,yang2014embedding,toutanova2015representing,trouillon2016complex} heavily rely on the subsymbolic representation of entities and relations learned through maximization of a scoring objective function over valid factual triples. Thus, the current success of such deep models hinges primarily on the power of those subsymbolic continuous real-valued representations in encoding the similarity/relatedness of entities and relations. Recent attempts have focused on neural multi-hop reasoners \cite{neelakantan2015compositional,peng2015towards,das2016chains,weissenborn2016separating,shen2017reasonet} to equip the model to deal with more complex reasoning where multi-hop inference is required. More recently, a Neural Theorem Prover \cite{rocktaschel2017end} has been proposed in an attempt to take advantage of both symbolic and sub-symbolic reasoning.

Despite their success, the main restriction common to 
machine learning-based reasoners is that they are unable to recognize and generalize to analogous situations or tasks. This inherent limitation follows from both the representation functions used and the learning process. The major issue comes from the mere reliance of these models on the representation of entities learned during the training or in the pre-training phase stored in a lookup table. Consequently, these models have difficulty to deal with out-of-vocabulary entities. Although the small-scale out-of-vocabulary problem has been addressed in part in the natural language processing domain by taking advantage of character-level embedding \cite{ling2015finding}, learning embeddings on the fly by leveraging text descriptions or spelling \cite{bahdanau2017learning}, copy mechanism \cite{eric2017copy} or pointer networks \cite{raghu2018hierarchical}, still these solutions are insufficient for transferring purposes. An even greater source of concern is that reasoning in most of the above sub-symbolic approaches hinges more on the notion of similarity and geometric-based proximity of real-valued vectors (induction) as opposed to performing transitive reasoning (deduction) over them.

Inspired by these observations, we take a different approach
in this work by investigating the emulation of deductive symbolic reasoning using memory networks. Memory networks \cite{weston2014memory} are a class of learning models capable of conducting multiple computational steps over an  explicit memory component before returning an answer. They have been recently applied successfully to a range of natural language processing tasks such as question answering \cite{sukhbaatar2015end, hill2015goldilocks}, language modeling \cite{sukhbaatar2015end}, and dialogue tasks \cite{bordes2016learning,dodge2015evaluating}. They use memory to store the context or knowledge bases of facts explicitly and perform inferencing over it using multi-hops recurrent attention. End-to-end memory networks (MemN2N) \cite{sukhbaatar2015end} are a less-supervised, more general version of these networks, applicable to the settings where labeled supporting memories are not available. They are very similar to the original memory networks, except that the
supporting memory slots have not been pre-determined as labels for the model. 
More specifically, first, those memory inputs useful for finding the correct answers are retrieved through an attention mechanism and an output vector is calculated by obtaining the weighted sum of the memory output representations. To apply multi-hop attention over the memory before outputting the response, the above process gets repeated recursively K times by replacing the query vector with the summation of the query and output vector obtained from the previous step. Finally, the output vector and final query representation from the last hop pass through a final weight matrix multiplication and a softmax to produce the output label. We have selected such networks since we believe that they are a primary candidate to perform well for deductive logical entailment. Their sequential nature corresponds, conceptually, to the sequential process underlying some deductive reasoning algorithms. The attention modeling corresponds to pulling only relevant information (logical axioms) necessary for the next reasoning step. And their success in natural language inferencing is also promising: while natural language inferencing does not follow a formal logical semantics, logical deductive entailment is nevertheless akin to some aspects of natural language reasoning. Besides, as attention can be traced over the run of a memory network, we will furthermore get insights into the "reasoning" underlying the network output, as we will be able to see which pieces of the memory (i.e., the input knowledge graph) are taken into account at each step. 

This paper contributes a recipe involving a simple but effective knowledge base triple normalization before learning their representation within an end-to-end memory network. To perform logical inference in more abstract level, and thereby facilitating the transfer of reasoning expertise from one knowledge graph to another, the normalization maps 
entities and predicates in a knowledge to a generic
vocabulary. Facts in additional knowledge bases are 
normalized using the same vocabulary, so that the 
network does not learn to overfit its learning to 
entity and predicate names in a specific knowledge
base. This emulates symbolic reasoning by neural embeddings as the actual names (as strings) of entities from the underlying logic such as variables, constants, functions, and predicates are insubstantial for logical entailment in the sense that a consistent renaming across a theory does not change the set of entailed formulas (under the same renaming). Thanks to the term-agnostic feature of our representation, we are able to create a reasoning system capable of performing reasoning over an unseen set of vocabularies in the test phase. 

Our approach combines the best of two worlds: transferability of classical deductive symbolic reasoning and robustness of neural-subsymbolic reasoning. This combination supports cross-knowledge graph reasoning obviating the need for supervised retraining over the task of interest or unsupervised pretraining over the external source of data for learning the representations when encountered with a new knowledge graph. 

Our contributions are threefold: (i) We present the construction of memory networks for emulating the symbolic deductive reasoning.  (ii) We propose an optimization to this architecture using normalization approach to enhance their transfer capability. We show that in an unnormalized setting, they fail to perform well across knowledge graphs.  (iii) We examine the efficacy of our model for cross-domain and cross-knowledge graph deductive reasoning. We also show the robustness of our model to the noisy train/test set and provides scalability (in terms of reduced time and space complexity) for large datasets.

This paper is structured as follows. In Section ~\ref{2} we discuss related research efforts, including a briefing of the history of attempts to integrate logical reasoning in connectionist approaches. In Section ~\ref{3} and ~\ref{4}, we concretely present the deep learning architecture we used. In Section ~\ref{5} and ~\ref{6}, we present an experimental evaluation of our approach and analyze our findings. We conclude and discuss future work in Section ~\ref{7}.

\section{Related works} \label{2}
The research into methods how to use artificial neural networks to perform logical deductive reasoning tasks is often referred to as the study of neural-symbolic integration. It can be traced back at least to the landmark 1942 article by McCulloch and Pitts \cite{mcculloch1943logical} in which it was shown how propositional logic formulas can be represented using a simple neural network model with threshold activation functions. A comprehensive and recent state of the art survey can be found in \cite{besold2017neural}, and hence we will only mention essentials for understanding the context our work is placed in.

Most of the body of work on neural-symbolic integration concerns propositional logic only (see, e.g.,\cite{garcez2008neural}), and indeed relationships both theoretical and practical in nature between propositional logics and subsymbolic systems are relatively easy to come by, an observation to which John McCarthy referred as the "propositional fixation" of artificial neural networks \cite{mccarthy1988epistemological}. Some of these include Knowledge-Based Artificial Neural Networks \cite{towell1994knowledge} and the closely related propositional core method \cite{hitzler2004logic,hoelldobler1994massiv}. Early attempts to go beyond propositional logic included the SHRUTI system \cite{shastri1999advances,shastri2007shruti} which, however, uses a non-standard connectionist architecture and thus had severe limitations as far as learning was concerned. Approaches to using standard artificial neural network architectures with proven good learning capabilities for first-order predicate logic \cite{gust2007learning} or first-order logic programming \cite{bader2008connectionist,bader2007fully} were by their very design unable to scale beyond toy examples.

In the past few years, deep learning as a subsymbolic machine learning paradigm has surpassed expectations as to the speed of progress in machine-learning based problem solving, and it is a reasonable assumption that these developments have not yet met their natural limit. Consequently, they are being looked upon as promising for trying to overcome the symbolic-subsymbolic divide \cite{asai2017classical,donadello2017logic,hohenecker2018ontology,maknideep,rocktaschel2017end,serafini2016learning,serafini2016logic} -- this list is not exhaustive. Even more work exists on inductive logical inference, e.g. \cite{nguyenconvolutional,rocktaschel2017end}, which is not what we are going to deal with in this work. Concretely, on the issue of doing logical reasoning using deep networks, we want to mention the following selected contributions:
Tensor-based approaches have been used \cite{rocktaschel2017end,serafini2016learning,serafini2016logic}, following \cite{grefenstette2015learning,socher2013reasoning}. However, approaches are restricted in terms of logical expressibility and/or to toy examples and limited evaluations.
\cite{maknideep} perform knowledge graph reasoning using RDF(S) \cite{world2014rdf,hitzler2009foundations}, based on knowledge graph embeddings.
However evaluation and training are done on the same knowledge graph, i.e., there is no learning of the general logical deduction calculus, and consequently no transfer thereof to new data.
\cite{hohenecker2018ontology} considers OWL RL reasoning \cite{hitzler2009foundations,hitzler2009owl}, however again training and evaluation are done on the same knowledge bases, i.e., no transfer is possible and no general deduction calculus is acquired during training. 

In short, to the best of our knowledge, to date, there is no sub-symbolic reasoning work, which is able to transfer the learning capability from one knowledge graph to unseen one. In fact, since previous works have focused to conduct reasoning on the unseen part of the same knowledge graph, they have tried to gain generalization ability through induction and robustness to missing edges\cite{guu2015traversing} as opposed to deduction. The induction queries include those triples (s,p,o) where at least there are one missing link in all the paths from s to o in the knowledge graph. Likewise, recent years have seen some progress in zero-shot relation learning in sub-symbolic reasoning domain\cite{neelakantan2015compositional, xiong2017deeppath, rocktaschel2015injecting}. Zero-shot learning refers to the ability of the model to infer new relations of pair of entities where that relation has not been seen before in training set\cite{bordes2011learning}. This generalization capability is still quite limited and fundamentally different from our work in terms of both methodology and purpose.

\section{Knowledge Graph Reasoning
}\label{3}

In order to explain what we are setting out to do, let us first re-frame the deductive reasoning (or entailment) problem as a classification task. Any given logic $\mathcal{L}$ comes with an entailment relation $\mathord{\models}_\mathcal{L} \sqsubseteq T_\mathcal{L}\times F_\mathcal{L}$, where $F_\mathcal{L}$ is a subset of the set of all logical formulas (or axioms) over $\mathcal{L}$, and $T_\mathcal{L}$ is the set of all theories (or sets of logical formulas) over $\mathcal{L}$. If $T\models\mathcal{L} F$, then we say that $F$ is \emph{entailed} by $F$. Re-framed as a classification task, we can ask whether a given pair $(T,F)\in T_\mathcal{L}\times F_\mathcal{L}$ should be classified as a valid entailment (i.e., $T\models_\mathcal{L} F$) holds, or as the opposite (i.e., $T\not\models_\mathcal{L} F$). Applying a deep learning approach to this, we would like to train a Deep Neural Network (DNN) on sets of examples $(T,F)$, such that the DNN learns to correctly classify them as valid or invalid inferences. Of course, we would have to restrict our attention to finite theories, which is usually done in computational logic anyway.

\subsection{Problem: Lack of Transferability}
We wish to train a model whose learnings will transfer to new theories within the same logic. That way, our results will demonstrate that the reasoning principles (inference rules) which underlie the logic have been learned. If we were to train a model such that it learns only to reason over one theory, then that could hardly be demonstrated.
One of the key obstacles we face with our task, however, is to understand how to represent training and test data so that they can be used in standard deep learning settings. Logical theories are highly structured, and it is essentially this structure which determines logical entailments, indeed some entailment algorithms can be understood in a straightforward way as a type of syntax rewriting systems. At the same time, the actual names (as strings) of entities from the underlying logic such as variables, constants, functions, predicates, are insubstantial for logical entailment in the sense that a consistent renaming across a theory does not change the set of entailed formulas (under the same renaming). 

For use with standard deep learning approaches, formulas -- or even theories -- will have to be represented in the real coordinate space $\mathbb{R}$ as vectors (points), matrices or tensors (multidimensional arrays); in deep learning, such a representation is commonly called an embedding. A plethora of embeddings for knowledge graphs have been proposed \cite{bordes2013translating,lin2015learning, trouillon2016complex,wang2014knowledge,yang2014embedding}, however, we are not aware of an existing embedding which adheres to the principles which seem important for the deductive reasoning scenario. Indeed, prominent use case explored for knowledge graph embeddings is not deductive in nature; rather, it concerns the problem of the discovery or suggestion of additional links or edges in the graph, together with appropriate edge labels. In this link discovery setting, the actual labels for nodes or edges in the graph, and as such their commonsense meanings, are likely important, and most existing embeddings reflect this. However, for deductive reasoning the names of entities are insubstantial, i.e., ideally should not be captured by an embedding. Another inherent problem in the use of such representations across knowledge graphs is the out-of-vocabulary problem. Formally speaking, such methods define a matrix $P \in \mathbb{R}^{d*|V|}$ to store a d dimensional real-valued vector of each word in the vocabulary $V$. Given the word lookup table $P$, the embedding for word $w$ can be obtained through multiplication of word lookup table $P$ and $w$'s one-hot vector representation ($1_{w}$) as $e_{w} = P.1_{w}$. The word lookup table can be initialized with vectors in an unsupervised task or during training of the reasoner. In any case, it is obvious that word lookup tables cannot generate vector representation for unseen terms and it is impractical to store the vectors of all words when vocabulary size is huge \cite{ling2015finding}.
 On the other hand, using standard graph embeddings \cite{cai2018comprehensive} also appears to be insufficient, because structural aspects such as the particular importance of nodes or edge labels from the RDF/RDFS (Resource Description Framework Schema) namespaces to the deductive reasoning process, would not get sufficient attention. 
Similarly, memory networks usually rely on word-level embeddings lookup tables, i.e., learned with the underlying rationale that words that occur in similar supervised scenarios should be represented by similar vectors in the real coordinate space. That is why they are known to have difficulties dealing with out-of-vocabulary terms, as a word lookup table cannot provide a representation for the unseen, and thus cannot really be applied to natural language inference over new sets of words \cite{bahdanau2017learning}, and for us this will pose a challenge in the transfer to new knowledge bases. 

We thus seek embeddings which are agnostic to the terms (i.e., strings) used as primitives in the knowledge base. One option may be to pursue variants of the copy mechanism and of pointer networks \cite{madotto2018mem2seq,raghu2018hierarchical} to refer to the unknown words in the memory in generating the responses. Despite the success of these mentioned methods in handling few unknown words absent during training, transferability and the ability of these models to generalize to a completely new set of vocabulary is still a widely open research question. Furthermore, these approaches are fundamentally appropriate for generative settings and therefore are not suitable for classification problems. Another option is utilizing character-level embeddings (ideal for open-vocabulary word representation) \cite{ling2015finding} to compose the representation of characters into words. Similarly, using character-level embeddings is an inelegant solution in our case, because of the importance of having a word-agnostic embedding. Therefore entity representation limitations of memory networks need to be overcome, in order to make them applicable to deductive logical entailment.

\subsection{Solution: Normalized Embedding
}
To build such an embedding, we will build on existing approaches for the embedding of structured data, and modify them for our purposes. We expect that some type of normalization will be required before embedding, and that this normalization will have two different aspects. One the one hand, normalization as usually done before invoking logical reasoning algorithms will help control the structural complexity of the formulas which constitute theories and entailments. On the other hand, we will explore syntactic normalization, and with this we mean a renaming of primitives from the logical language (variables, constants, functions, predicates) to a set of predefined entity names which will be used across different theories. By 
randomly assigning the mapping for the renaming, the network's learning will be based on the structural information within the theories, and not on the actual names of the primitives, which should be insubstantial for the entailment task.
Note that the normalization does not only play the role of ``forgetting'' irrelevant label names, but also
makes it possible to transfer learning from one knowledge graph to the other.
Indeed, for the approach to work, the network should be trained with many knowledge graphs, and then subsequently tested on completely new ones which had not been encountered during training. 
Our preliminary results show how our simple but very effective normalization phase can surprisingly lead to creating a word-agnostic reasoning system capable of conducting reasoning over unseen knowledge graphs containing new vocabulary.

\section{Model Architecture} \label{4}
Our model architecture is an adaptation of the end-to-end memory network proposed by \cite{sukhbaatar2015end} with some fundamental alterations necessary for abstract reasoning. A high-level view of our model shown in Figure~\ref{fig:model} is as follows. It takes a discrete set of normalized triples $t_{1}, ..., t_{n}$ that are to be stored in the memory, a query $q$, and outputs a "yes" or "no" as an answer to determine whether $q$ can be inferred from the current knowledge graph statements or not. Each of the normalized $t_{i}$ and $q$ contains symbols coming from a general dictionary with $V$ normalized words shared among all of the knowledge graphs in both training and test sets. The model writes all triples to the memory and then calculates a continuous embedding for the whole triples and $q$. Through multiple hop attention over those continuous representations, the model then classifies the query. The model will get trained by back-propagation of error from output to the input through multiple memory accesses and embeddings will get learned through these accesses. We discuss 
components of the architecture in more detail below.

\subsection{Model Description}
\begin{figure*}[!t]

   \includegraphics[width=.95\textwidth]{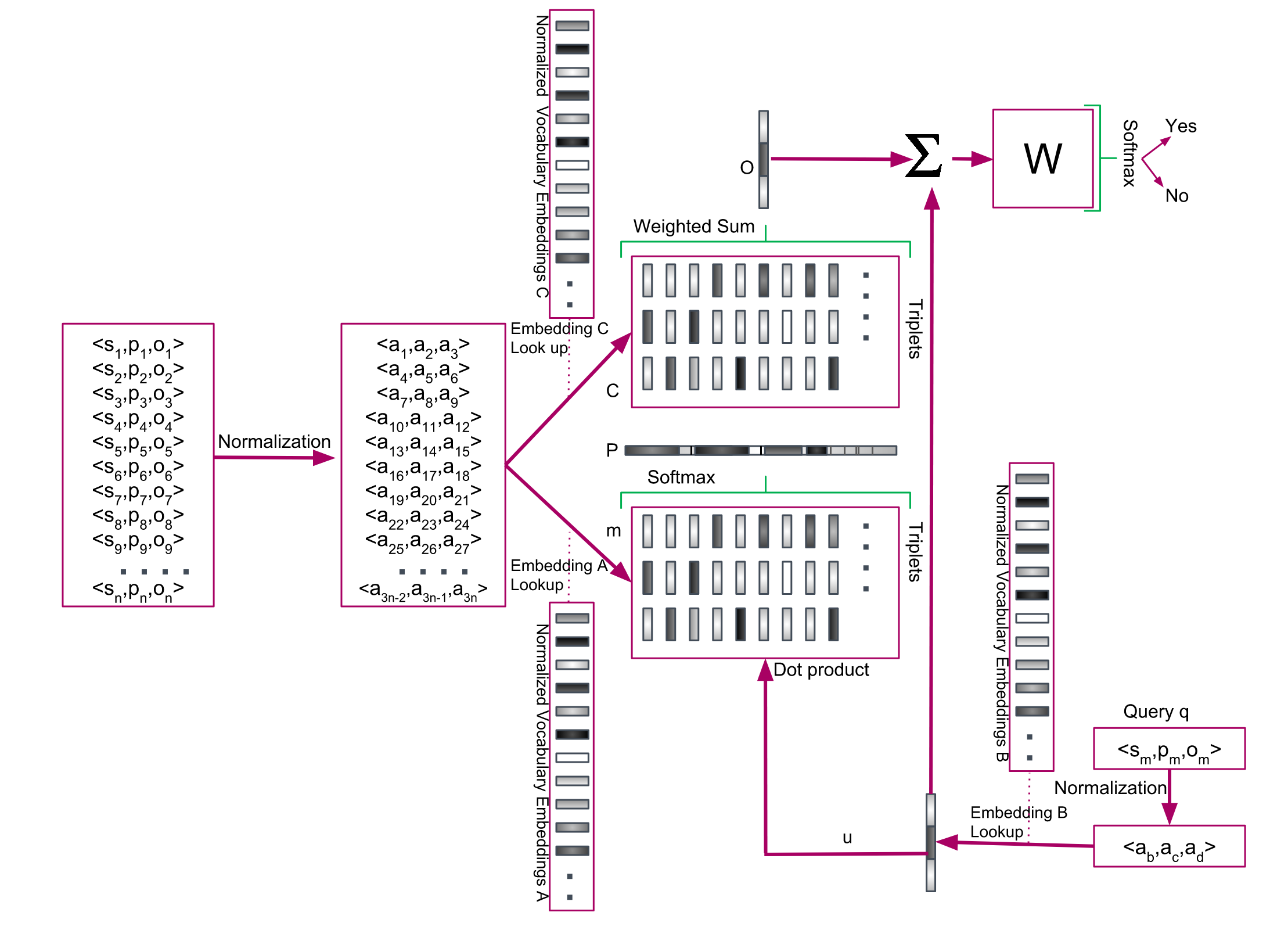}
   \caption{Proposed model diagram for K=1}
   \label{fig:model}
\end{figure*}
The design of the model is based on the MemN2N\cite{sukhbaatar2015end} end-to-end memory network. The model is augmented with 
an external memory component storing the embedding of the normalized triples in our knowledge graph. This external memory is defined as an $n\times 3 \times d$ tensor where $n$ denotes the number of triples in knowledge graph and $d$ is the dimensionality of the embeddings. The knowledge base is stored in the memory vectors from two continuous representations of $m_{i}$ and $c_{i}$ obtained from two input and output embedding matrices of A and C with size  $d \times V$ where $V$ is the size of vocabulary. Similarly, the query $q$ is embedded via a matrix $B$ to obtain an internal state $u$.  
In each reasoning step, those memory slots useful for finding the correct answers should have their contents retrieved. To enable
this, we use an attention mechanism for $q$ over memory input representations by taking 
an internal product followed by a softmax:
\begin{equation}
p_i = Softmax(u^T(m_{i}))
\label{eq:pi}
\end{equation}
where
$$Softmax(a_{i}) = \frac{e^{(a_{i})}}{\sum_{j}e^{(a_{j})}}$$
Equation~\ref{eq:pi} is used to calculate a probability vector $p$ over the memory inputs, the output vector $o$ is computed as weighted sum of the transformed memory contents $c_{i}$s with respect to their corresponding probabilities $p_{i}$ by:
\begin{equation}o=\sum_{i}p_{i}c_{i}\end{equation}
This describes the computation within a single hop. The internal state of query vector gets updated for the next hop using: 

\begin{equation}u^{k+1}=u^{k}+o^{k}\end{equation}
The process repeats $K$ times where $K$ is the number of memory units in the network. The output of the $K^{th}$
memory unit is used to predict a label by passing $o^K$ and $u^K$ through a weight matrix of size $V \times d$ and a softmax:
\begin{equation}\hat{a}=Softmax(W(u^{K+1}))=Softmax(W(u^{k}+o^{k}))\end{equation}
Figure~\ref{fig:model} illustrates the model for $K=1$. The parameters to be learned 
by backpropagation are the matrices $A, B, C,$ and $W$.

\subsection{Memory Content}

A knowledge graph $(\epsilon, R)$ is a collection of facts stored as triplets $(e1,r,e2)$ where $e1
 \in \epsilon$ and $e2 \in \epsilon$ are subject and object respectively, while $r \in R$ is a predicate (relation) binding $e1$ and $e2$ together. Every entity in the knowledge graph is represented by a unique Universal Resource Identifier (URI). We normalize these triples by systematically renaming all URIs which are not in the RDF/RDFS namespaces as discussed previously. Each such URI is mapped to a set of arbitrary strings in a predefined set $\mathcal{A} = \{a_1, . . . , a_n\}$, where $n$ is taken as a training hyper-parameter giving an upper bound for the largest number of entities in a knowledge graph the system will be able to handle.
Note that URIs in the RDF/RDFS namespaces are not renamed, as they are important for the deductive reasoning process. 
Consequently, each normalized knowledge graph will be a collection of facts stored as triplets $\{(a_i,a_j,a_k)\}$. 

It is important to note that each symbol is mapped into an element of $\mathcal{A}$ regardless of its position in the triple, and whether it is a subject or an object or a predicate. Yet the position of a subject within a predicate is an important feature to consider. 
Inspired by \cite{sukhbaatar2015end}, we thus employ a positional encoding to encode the position of each element within the triplet. 
This gives the formal solution: $m_{i} = \sum_{j}{l_{j}}\cdot Ax_{ij}$ , where $\cdot$ denotes an element-wise multiplication and $l_{j}$ is a column vector with the structure $l_{kj} = (1 - j/3) - (k/d)(1-2j/3)$ (assuming 1-based indexing), with denominator of $3$ (the number of elements in the triplet), and $d$ is the size of the embedding vector in the memory input embedding matrix $A$. Each memory slot thus represents the positional-weighted summation of each triplet. By using positional encoding (PE), we ensure that the order of the elements now affects the encoding of each memory slot $m_{i}$. This representation which is used for query triplet, memory inputs, and memory outputs, is core to everything we do.

\section{Experimental Setups}\label{5}
\begin{table*}[]
\begin{adjustbox}{width=1\textwidth}
\small
\begin{tabular}{|l|l|}
\hline
Dataset & \multicolumn{1}{c|}{Ontologies} \\ \hline
\multicolumn{1}{|c|}{OWL-Centric} & \begin{tabular}[c]{@{}l@{}}Amino Acid Ontology schema, Biological Pathway Exchange (BioPAX) schema, COmmon Semantic MOdel (COSMO),\\ dbpedia-schema, Descriptions and situation, Disease, Dolce, Dublin\_core schema, Gene, General formal ontology (GFO),\\ Human Phenotype, Institutional Ontology, Metadata for Ontology Description and publication, \\Ontology for Biomedical Investigations, Phenotypic quality, Schema.org, University of Lehigh benchmark, \\Xenopus anatomy and development, Yet Another More Advanced Top-level Ontology (YAMATO).\\ \end{tabular} \\ \hline
\multicolumn{1}{|c|}{Linked Data} & \begin{tabular}[c]{@{}l@{}}AGROVOC Linked Dataset, Amsterdam Museum Linked Open Data, The Apertium Bilingual Dictio- naries on the Web of\\ Data, A Curated and Evolving Linguistic Linked Dataset (Asit), EARTh: an Environmental Application Reference Thesaurus\\ in the Linked Open Data Cloud data, lemonUby - a large, interlinked, syntactically-rich lexical resource for ontologies, \\Linked European Television Heritage data, Linked Web APIs Dataset: Web APIs meet Linked Data.\end{tabular} \\ \hline
\multicolumn{1}{|c|}{OWL-Centric Test Set} & \begin{tabular}[c]{@{}l@{}}Animal Health Surveillance Ontology, Cryptographic ontology of Semantic interoperability for rapid integration and deployment,\\ Drug Abuse Ontology, Drug target ontology, General Ontology for Linguistic Description (GOLD), Identification ontology,\\ Inline Hockey League pattern ontology, Knowledge processing ontology for Robots, Minimal category of list ontology, \\Provenance and Plans ontology , SAREF: the Smart Appliances REFerence ontology,\\ Tatian Corpus of Deviating Examples (T-CODEX) ontology\end{tabular} \\ \hline

\end{tabular}
\end{adjustbox}
\caption{List of ontologies used to create our datasets}
\label{my-label}
\end{table*}

\begin{table*}[]
\begin{adjustbox}{width=1\textwidth}
\begin{tabular}{|l|l|llllll|llllll|l|}
\hline
\multicolumn{1}{|c|}{\multirow{2}{*}{Test Dataset}} & \multirow{2}{*}{\#KG} & \multicolumn{6}{c|}{Base}                                                                                                                                & \multicolumn{6}{c|}{Inferred}                                                                                                                            & \multicolumn{1}{c|}{Invalid} \\ \cline{3-15} 
\multicolumn{1}{|c|}{}                              &                      & \multicolumn{1}{l|}{\#Facts} & \multicolumn{1}{l|}{\#Ent.} & \multicolumn{1}{l|}{\%Class} & \multicolumn{1}{l|}{\%Indv} & \multicolumn{1}{l|}{\%R.} & \%Axiom. & \multicolumn{1}{l|}{\#Facts} & \multicolumn{1}{l|}{\#Ent.} & \multicolumn{1}{l|}{\%Class} & \multicolumn{1}{l|}{\%Indv} & \multicolumn{1}{l|}{\%R.} & \%Axiom. & \#Facts                       \\ \hline
OWL-Centric                                         & 2464                 & 996                         & 832                        & 14                          & 19                         & 3                        & 0       & 494                         & 832                        & 14                          & 0.01                       & 1                        & 20      & 462                          \\ \hline
Linked Data                                         & 20527                & 999                         & 787                        & 3                           & 22                         & 5                        & 0       & 124                         & 787                        & 3                           & 0.006                      & 1                        & 85      & 124                          \\ \hline
OWL-Centric Test Set                                & 21                   & 622                         & 400                        & 36                          & 41                         & 3                        & 0       & 837                         & 400                        & 36                          & 3                          & 1                        & 12      & 476                          \\ \hline
Synthetic Data                                      & 2                    & 752                         & 506                        & 52                          & 0                          & 1                        & 0       & 126356                      & 506                        & 52                          & 0                          & 1                        & 0.07    & 700                          \\ \hline
\end{tabular}
\end{adjustbox}
\caption{Statistics of various datasets used in experiments}
\label{my-label}
\end{table*}
\subsection{Candidate Logic}
There is a plethora of logics which could be used for our investigation. We exclude propositional logics because they seem to be easier to capture using connectionist architectures, while at the same time the methods used for dealing with them in a subsymbolic manner do mostly not seem to transfer to non-propositional logics. Here we use Resource Description Framework (RDFS). The Resource Description Framework (RDF) \cite{world2014rdf,hitzler2009foundations} is an established and widely used W3C standard for expressing knowledge graphs. The standard comes with a formal semantics which defines an entailment relation. As a logic, RDFS is of very low expressivity, and reasoning algorithms are very straightforward. One way to frame it is that there is a small set of thirteen entailment rules, fixed across all knowledge graphs, which are expressible using Datalog. These thirteen rules can be used to entail new facts. The completion of a knowledge graph is in general infinite because, by definition, there is an infinite set of facts (related to RDFS-encodings of lists) which are always entailed - however for practical reasons, this is ignored by established RDFS reasoning systems, i.e., for all practical purposes we can consider completions of knowledge graphs to be finite.
\subsection{Dataset}

Due to the novel nature of the problem at hand, there is no dataset available for testing the capability of our approach. The good news however is, there are plethora of knowledge graph\cite{cheatham2018geolink} that we could use to create our own dataset. The Linked Data Cloud\footnote{https://lod-cloud.net/} website lists over 1,200 interlinked RDF(S) datasets, which constitute knowledge graphs suitable for our setting, some of which are of substantial size. Here, we have collected data from this website as well as Data Hub\footnote{https://datahub.io/} website to create our training set\footnote{https://github.com/md-k-sarker/KG-Cmpl-dataset}. Our training set (owl-centric dataset) is comprised of a set of knowledge graphs of size 1000 triples sampled from around 20 ontologies (as listed in Table 1). In order to test our model generalization ability to completely different domain, we have collected another dataset called OWL-Centric test set. Furthermore, to assure our evaluation set represents the real-world RDF data and meet the quality requirement of linked data\cite{janowicz2014five}, we have followed \cite{sam2018quality} to collect data for our Linked Data dataset. To our best attempt we could not find public rdf-dump or sparql endpoint for some of the datasets mentioned in the paper though. In addition, to test the capability of our model to conduct long chain of reasonings we have created a synthetic dataset using rdfs:subclass and rdfs:subproperty predicates. It covers the reasoning chains of maximum 10.

For each knowledge graph we have created a finite set of inferred triples using Apache Jena\footnote{https://jena.apache.org/} api. These inferred triples comprise our positive class instances. For generating invalid instances we are following two methods. In the first scenario, we are generating invalid triplets by random permutation of entities and filtering out those triplets which are currently in the knowledge base or set of valid triplets. In the second scenario, which serves as the final quality check for not including trivial invalid triplets in our dataset, we are creating invalid instances with the aid of rdf:type predicate. More specifically, for each valid triple in the dataset, we replace one of the elements (chosen randomly), with another random element which qualifies for being placed in that triple based on its type-of relationships. Here, as well, we add the newly generated triplet if it is not already part of original knowledge base or valid facts. Indeed, through random selection of one of the hyponyms of hypernyms of entity of interest, we assure that our created dataset is challenging enough. Those datasets created by this strategy are denoted by superscript "a" in experimental Table 3. Some important statistics of our created datasets has been summarized in Table 1 and 2. More specifically, Table 2 illustrates number of knowledge graphs in each of our datasets and average number of facts and entities per knowledge graph. It also demonstrates average number of classes, individuals, relations and axiomatic triples for each knowledge graph (in percentage).
\subsection{Training Details}
Trainings have been done over 10 epochs using Adam optimizer. All trainings have been done with the batch size of 100 over triplets. For the final batches of queries for each knowledge graph, we have used zero-padding to the maximum batch size of 100. The capacity of our external memory is 1000 which is the maximum size of our knowledge bases also. Our model have been trained using a learning rate of $\eta = 0.005$ and learning rate decay of $\eta/2$. We have used linear starting of 1 epoch where we have removed the softmax from each memory layer except for the final layer. L2 norm clipping of max 40 has been applied on gradient. Both memory input embeddings and memory output embeddings are vectors of size $20\times1$. The embedding matrices of A, B, and C therefore are of size $|V|\times d = 3033 \times 20$ where 3033 is the size of normalized vocabulary. Unless otherwise mentioned, we have used K=10 for all of our experiments. Adjacent weight sharing has been used where output embedding of one layer is the input embedding of the next one as in $A^{k+1} = C$. Similarly, the answer prediction weight matrix $W$ get copied to the final output embedding $C^K$ and query embedding is equal to the first layer input embedding as in $B = A^1$. All the weights are initialized using Gaussian distribution with $\mu = 0$ and $\sigma = 0.1$.
\section{Experimental Results}\label{6}
\begin{table*}[]
\begin{adjustbox}{width=1\textwidth}
\begin{threeparttable}
\begin{tabular}{|l|l|l|l|l|l|l|l|l|l|}
\hline
\multicolumn{1}{|c|}{\multirow{2}{*}{Training Dataset}} & \multirow{2}{*}{Test Dataset}                                                                           & \multicolumn{3}{c|}{Valid Triples Class} & \multicolumn{3}{c|}{Invalid Triples Class} &\multicolumn{1}{|c|}{\multirow{2}{*}{Accuracy}}\\ \cline{3-8} 
\multicolumn{1}{|c|}{}                                  &                                                                                                         & Precision     & {\begin{tabular}[c]{@{}c@{}}Recall\\ /Sensitivity\end{tabular}}   & F-measure    & Precision     & {\begin{tabular}[c]{@{}c@{}}Recall\\ /Specificity\end{tabular}}    & F-measure  &  \\ \hline
OWL-Centric Dataset                                  & Linked Data                                                                        & 93            & 98        & 96           & 98            & 93         & 95   &\textbf{96}         \\
\multicolumn{1}{|c|}{OWL-Centric Dataset (90\%)}        & OWL-Centric Dataset (10\%)                                                                              & 88            & 91        & 89           & 90            & 88         & 89&\textbf{90}            \\
OWL-Centric Dataset                                     & OWL-Centric Test Set \tnote{b}                                                                  & 79            & 62        & 68           & 70            & 84         & 76 &\textbf{69}           \\
OWL-Centric Dataset                                     & Synthetic Data                                                                                          & 65            & 49        & 40           & 52            & 54         & 42 &\textbf{52}           \\ \hline

OWL-Centric Dataset                                     & Linked Data \tnote{a}                                                                      & 54           & 98        & 70           & 91            & 16         & 27 &86           \\
OWL-Centric Dataset \tnote{a}                                     & \begin{tabular}[c]{@{}l@{}}  Linked Data \tnote{a} \end{tabular} & 62            & 72        & 67           & 67            & 56         & 61    &91        \\ 
OWL-Centric Dataset(90\%) \tnote{a}                                     & \begin{tabular}[c]{@{}l@{}}  OWL-Centric Dataset(10\%) \tnote{a} \end{tabular} & 79            & 72        & 75           & 74            & 81         & 77 &80           \\ 

OWL-Centric Dataset                                     & \begin{tabular}[c]{@{}l@{}}OWL-Centric Test Set  \tnote{a} \thinspace \tnote{b}\end{tabular} & 58            & 68        & 62           & 62            & 50         & 54 &58           \\ 

OWL-Centric Dataset \tnote{a}                                   & \begin{tabular}[c]{@{}l@{}}OWL-Centric Test Set  \tnote{a} \thinspace \tnote{b}\end{tabular} & 77            & 57        & 65           & 66            & 82         & 73   &73        \\
OWL-Centric Dataset                                     & Synthetic Data \tnote{a}                                                                    & 70            & 51        & 40           & 47            & 52         & 38 &51           \\

OWL-Centric Dataset \tnote{a}                                     & \begin{tabular}[c]{@{}l@{}}  Synthetic Data \tnote{a} \end{tabular} & 67            & 23       & 25           & 52            & 80         & 62   &50         \\
 \hline\hline
\multicolumn{9}{|c|}{\textbf{Baseline}}\\ \hline
OWL-Centric Dataset                                  & Linked Data                                                                        & 73            & 98        & 83           & 94           & 46          & 61     &43       \\
\multicolumn{1}{|c|}{OWL-Centric Dataset (90\%)}        & OWL-Centric Dataset (10\%)                                                                              & 84            & 83        & 84           & 84            & 84         & 84  &82          \\
OWL-Centric Dataset                                     & OWL-Centric Test Set \tnote{b}                                                                  & 62            &84         &70            &80             & 40         & 48 &61           \\
OWL-Centric Dataset                                     & Synthetic Data                                                                                          & 35            & 41       & 32           & 48            & 55       & 45 &48             \\

\hline

\hline
\end{tabular}

        \begin{tablenotes}
            \item[a] More Tricky Nos \& Balanced Dataset
            \item[b] Completely Different Domain.
        \end{tablenotes}
       
        \caption{Experimental results of proposed model}
\label{my-label}
\end{threeparttable}
 \end{adjustbox}
\end{table*}
\subsection{Quantitative Results}

In this section, we highlight some of our findings along with experimental results of our proposed approach. The evaluation metrics we report are average of precision and recall and f-measure over all the knowledge graphs in the test set, obtained for both valid and invalid set of triplets. Particularly, we also report, the recall for negative class, also called specificity, to interpret the result more carefully by counting number of true negatives. Additionally, as we mentioned in the training detail, we have done zero-padding for each batch of query triplets of size of less than 100. This implies the need for introducing another class label for such zero paddings both in the training and test phase. In our evaluation, however, we have not considered the zero-padding class in calculation of precision, recall and f-measure. Through our evaluations, however, we have observed some misclassification from/to this class. Here we are reporting accuracy as well to take such mistakes also into account. 

To the best of our knowledge there is no architecture capable of conducting deductive reasoning on completely unseen knowledge graph. That is why, we have considered the non-normalized embedding version of our memory network as a baseline. Our technique shows a clear significant advantage over the baseline as shown in Table 3. A further even more important benefit of using our normalization model is its training time. In fact, this considerable time complexity difference is the result of remarkable size difference of embedding matrices in original and normalized cases. For instance, the size of embedding matrices to be learned by our algorithm for normalized OWL-Centric dataset is $3033 \times 20$ as opposed to  $811261 \times 20$ for normalized one (and $1974062 \times 20$ for Linked Data which is prohibitively big). That has caused a remarkably high decrease in training time and space complexity and consequently has helped the scalability of our memory networks. In case of OWL-Centric dataset, for instance, the space required for saving normalized model is 80 times less than the intact model($\approx 4G$ after compression). Likewise, the normalized model is almost 40 times faster to train than the non-normalized one for this dataset. Our normalized model trained for just over a day on OWL-Centric data achieves better accuracy, whereas it trained on the same non-normalized dataset more than a week on 12-core machine. Hence, the importance of using our novel normalized representation learning cannot be emphasized too much.

To further get an idea of how our model performs on different data sources, we have applied our approach to multiple datasets with various characteristics. The result across all variations are given in Table 3. From this Table we can see that, apart from our strikingly good performance compared to the baseline, there are number of other interesting points: Our model gets even better results on Linked Data task while it has trained on the OWL-Centric dataset. The reasons for this performance gain are not yet wholly understood. Another interesting observation is the poor performance of our algorithm when it has trained on OWL-Centric dataset and has been applied on a tricky version of the Linked Data. In that case our model has classified most of the triples to the "yes" class and this has led to low specificity (recall for "no" class) of 16\%. It is inevitable because in challenging "No"'s version of our dataset the negative instances bear close resemblance to positives ones, making differentiation more challenging. Training the model on the tricky OWL-Centric dataset has improved that by substantial margin (more than three times).
In case of the synthetic data, however, although performance is not ideal , we still nevertheless believe that it is acceptable. An evident rationalization for this performance decrease for synthetic data compared to other sources of data is the significant difference in reasoning patterns and nature of training and test set. Indeed, our training so far have only been done on real-world datasets and not peculiar synthetic data. Training the model on synthetic data is not the focus of this study.
\begin{table*}[]
\centering
\begin{adjustbox}{width=1\textwidth}
\small
\begin{threeparttable}
\begin{tabular}{|l|l|l|l|l|l|l|l|l|l|l|l|l|l|l|l|l|l|l|l|l|l|l|l|l|l|l|l|l|l|l|l|l|l|}
\hline
\multicolumn{1}{|c|}{\multirow{2}{*}{Test Dataset}} & \multicolumn{3}{c|}{Hop 0} & \multicolumn{3}{c|}{Hop 1} & \multicolumn{3}{c|}{Hop 2} & \multicolumn{3}{c|}{Hop 3} & \multicolumn{3}{c|}{Hop 4} & \multicolumn{3}{c|}{Hop 5} & \multicolumn{3}{c|}{Hop 6} & \multicolumn{3}{c|}{Hop 7} & \multicolumn{3}{c|}{Hop 8} & \multicolumn{3}{c|}{Hop 9} & \multicolumn{3}{c|}{Hop 10} \\ \cline{2-34} 
\multicolumn{1}{|c|}{} & P & R & F & P & R & F & P & R & F & P & R & F & P & R & F & P & R & F & P & R & F & P & R & F & P & R & F & P & R & F & P & R & F \\ \hline
Linked Data\tnote{a} & 0 & 0 & 0 & 80 & 99 & 88 & 89 & 97 & 93 & 77 & 98 & 86 & - & - & - & - & - & - & - & - & - & - & - & - & - & - & - & - & - & - & - & - & - \\ \hline
Linked Data\tnote{b} & 2 & 0 & 0 & 82& 91 & 86& 89 & 98 & 93 & 79 & 100 & 88 & - & - & - & - & - & - & - & - & - & - & - & - & - & - & - & - & - & - & - & - & - \\ \hline
OWL-Centric \tnote{c} & 19 & 5 & 9 & 31 & 75 & 42 & 78 & 80 & 78 & 48 & 47 & 44 & 4 & 34 & 6 & - & - & - & - & - & - & - & - & - & - & - & - & - & - & - & - & - & - \\ \hline
Synthetic & 32 & 46 & 33 & 31 & 87 & 38 & 66 & 55 & 44 & 25 & 45 & 32 & 29 & 46 & 33 & 26 & 46 & 33 & 25 & 46 & 33 & 25 & 46 & 33 & 24 & 43 & 31 & 25 & 43 & 31 & 22 & 36 & 28 \\ \hline
\end{tabular}

\begin{tablenotes}
            \item[a] LemonUby Ontology
            \item[b] Agrovoc Ontology
            \item[c] Completely Different Domain
        \end{tablenotes}

\end{threeparttable}
\end{adjustbox}
\caption{Experimental results over each reasoning hop}
\label{my-label}
\end{table*}
\begin{table*}[]
\centering
\small
\begin{threeparttable}
\begin{tabular}{|l|l|l|l|l|l|l|l|l|l|l|}
\hline
\multicolumn{1}{|c|}{Dataset} & Hop 1 & Hop 2 & Hop 3 & Hop 4 & Hop 5 & Hop 6 & Hop 7 & Hop 8 & Hop 9 & Hop 10 \\ \hline
\textbf{\textit{OWL-Centric}} \tnote{a} & 8\% & 67\% & 24\% & 0.01\% & 0\% & 0\% & 0\% & 0\% & 0\% & 0\% \\ \hline
Linked Data\tnote{b} & 31\% & 50\% & 19\% & 0\% & 0\% & 0\% & 0\% & 0\% & 0\% & 0\% \\ \hline
Linked Data\tnote{c} & 34\% & 46\% & 20\% & 0\% & 0\% & 0\% & 0\% & 0\% & 0\% & 0\% \\ \hline
OWL-Centric\tnote{d} & 5\% & 64\% & 30\% & 1\% & 0\% & 0\% & 0\% & 0\% & 0\% & 0\% \\ \hline
Synthetic Data & 0.03\% & 1.42\% &  1\%& 1.56\% & 3.09\% & 6.03\% & 11.46\% & 20.48\% & 31.25\% & 23.65\% \\ \hline
\end{tabular}
\begin{tablenotes}
			\item[a] \textit{Training Set}
            \item[b] LemonUby Ontology
            \item[c] Agrovoc Ontology
            \item[d] Completely Different Domain
        \end{tablenotes}
\end{threeparttable}
\caption{Data distribution per knowledge graph over each reasoning hop}
\label{my-label}
\end{table*}

Further experiments were needed to analyze the reasoning depth acquired by our network. Fundamentally, we conjecture that reasoning depth acquired by the network will correspond both to (1) the number of layers in the deep network, and (2) the ratio of deep versus shallow reasoning required to perform the deductive reasoning. Let us explain this. Forward-chaining reasoners (which are the standard for RDF(S), OWL EL, and OWL RL reasoning, and can also be used for Datalog) iteratively apply inference rules in order to derive new entailed facts. In subsequent iterations, the previously derived facts need to be taken into account. The number of sequential applications of the inference rules which are required to obtain a given logical consequence can be understood as a measure of the "depth" of the deductive entailment. To gain better understanding of what our model has learned, we have mimic this behavior of symbolic reasoners in creating our test set. In order to do that, we first have started from our input knowledge graph $K_{0}$ in hop 0. We then have produced, subsequently, knowledge graphs of $K_{1}$, $K{_2}$, $K{_3}$ until no new triples are added (i.e. until $K_{n}$ is empty) by applying the RDFS inference patterns from W3C website\footnote{https://www.w3.org/TR/rdf11-mt/\#rdfs-entailment}\textsuperscript{,}\footnote{https://www.w3.org/TR/rdf11-mt/\#entailment-rules-informative}. Consequently, our hop 0 dataset contains the original graph, and the inferred axioms thus are replaced by the triples in the original graph. Hop 1 contains the RDFS axiomatic triples in the inferred axioms field. The real inference steps start with $K_{n}$ where $n>=2$. It is worthwhile noting that, in the process of creating this data, our reasoning tool encountered several errors during reasoning because of the missing entities in some triples. There are a lot of such a missing/unknown/inapplicable entities in real-world knowledge graphs emphasizing the need for using more robust sub-symbolic reasoners. 

Table 4 summarizes our results in this setup. The poor performance on hop 0 is not unexpected since our training set does not include any original triplets. Unsurprisingly, we also observe that result over our synthetic data generated with subclass-of and sub-property-of predicates is poor. That is because of the huge gap between distribution of our training data over reasoning hops and the synthetic data reasoning hop length distribution. Table 5 provides further evidence for that. From Tables 4 and 5, one can see how the distribution of our training set affects the learning capability of our model. Apart from our observations, previous studies\cite{das2017go,rocktaschel2017end,das2016chains,yang2017differentiable} also corroborate that the reasoning chain length required to answer a query in a real-world KB is limited to 3 or 4. Therefore a synthetic training toy set is required to be built for further analysis of the reasoning depth capability of our model in future.

Furthermore, a naive expectation on the trained network would be that each layer would perform a type of equivalent to an inference rule application. If this is the case, then the number of layers would limit the entailment depth the network could acquire, however we had to assess this assumption experimentally. Therefore, we have done 10 experiments (K=1 to 10) to assess the effect of change of number of computational hops on our results over OWL-Centric Dataset. Interestingly, our experimental results suggest that our model is able to get almost the same performance with K=1 and more interestingly, the F-measure remains constant with step by step increase of K from 1 to 10. This shows us that multi-hop reasoning can be done in one-hop attention of memory networks (as we need only 2-3 hop reasoning) over our training set, while the increase of number of hops would not hurt our performance. This demonstrates robustness of the proposed method against change of its structure. This also suggests that each attention hop of our memory network is able to conduct more than one inference rule application step (deductive reasoning hop).

\subsection{General Embeddings Visualization}
In order to gain some insight on the behavior of our normalized embedding model, we have plotted a t-Distributed Stochastic Neighbor Embedding (t-SNE) \cite{maaten2008visualizing} and Principal Component Analysis (PCA) two-dimensional vector visualization of embeddings computed for the RDF(S) words and all normalized words in the knowledge graphs in figures 2, 3, and 4 respectively. The embeddings have been fetched from the matrix B (embedding query lookup table) in the computational hop 1 of our trained model over OWL-Centric dataset. Words are positioned in the plot by the semantic relationships implied by their embeddings. As anticipated all the normalized words tend to form one cluster as opposed to creating multiple separated clusters as shown in Figure 3 and 4. We found PCA plot more insightful. The PCA projection illustrates ability of our model to automatically organize RDF(S) concepts and learn implicitly
the relationships between them, as during the training we have not provide any supervised information about what each RDF(S) element means. For instance, rdfs:domain and rdfs:range have been located very close together and far from normalized entities. Similarly, rdf:subject, rdf:predicate and rdf:object are very similar in the vector space. That is the case for rdfs:seesAlso and isDefinedBy. Similarly, rdfs:container, rdf:bag, rdf:seq, and rdf:alt are in vicinity of each other. Rdf:lanstring is the only RDF domain entity which is inside the normalized entities circle. We believe that it is because the rdf:langString's domain and range is string and consequently it has only co-occurred with normalized instance in knowledge graph. Thus its vector is very close to normalized entities vectors. Another possible rationalization for this might be the low frequency of rdf:langString in our training set. This could possibly contribute to the general representation for this word.
   
   

\begin{figure}
\centering
\begin{subfigure}{.5\textwidth}
  \centering
  \includegraphics[width=1\linewidth]{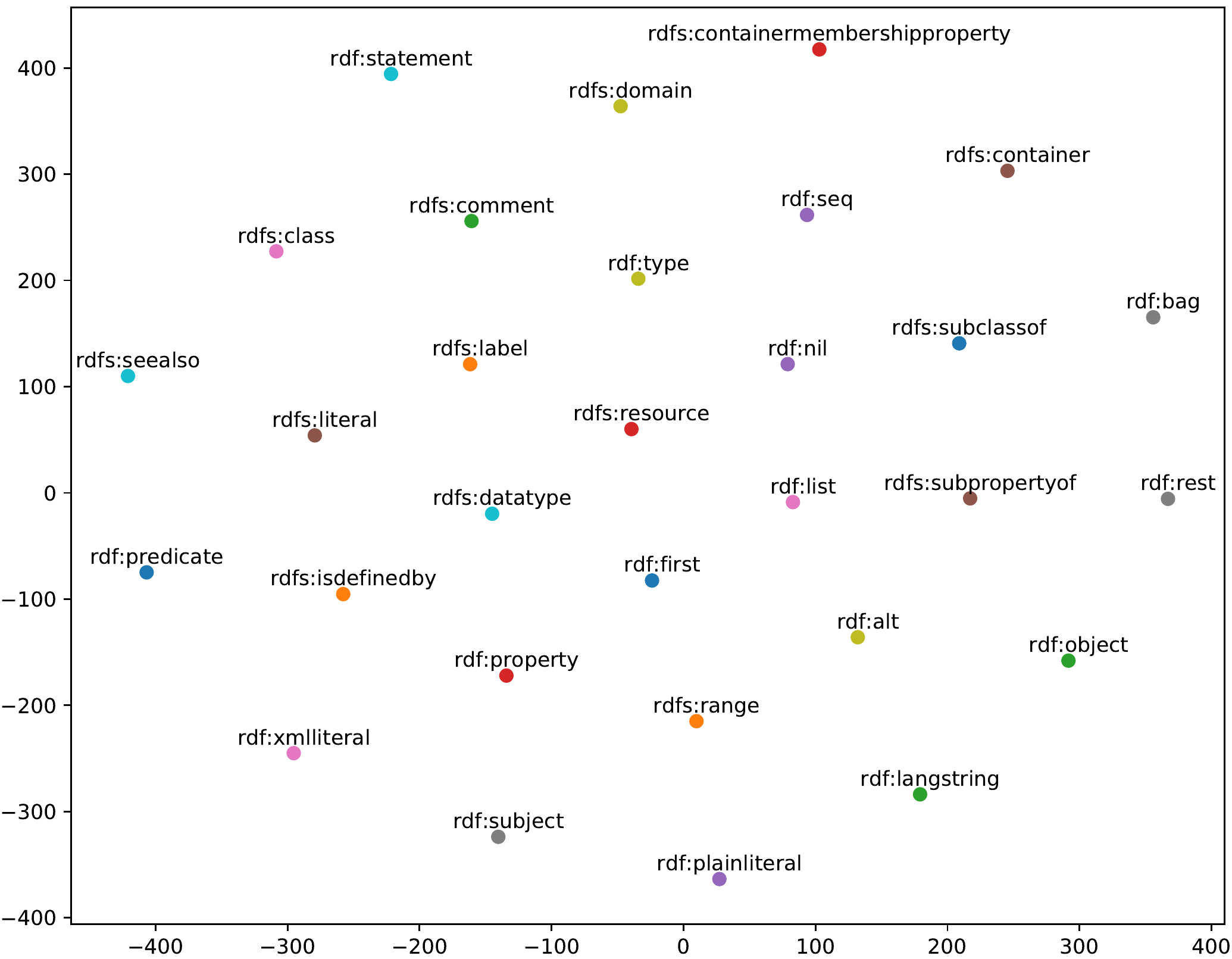}
  \caption{Embeddings for RDF(S) namespace}
  \label{fig:sub1}
\end{subfigure}%
\begin{subfigure}{.5\textwidth}
  \centering
  \includegraphics[width=1\linewidth]{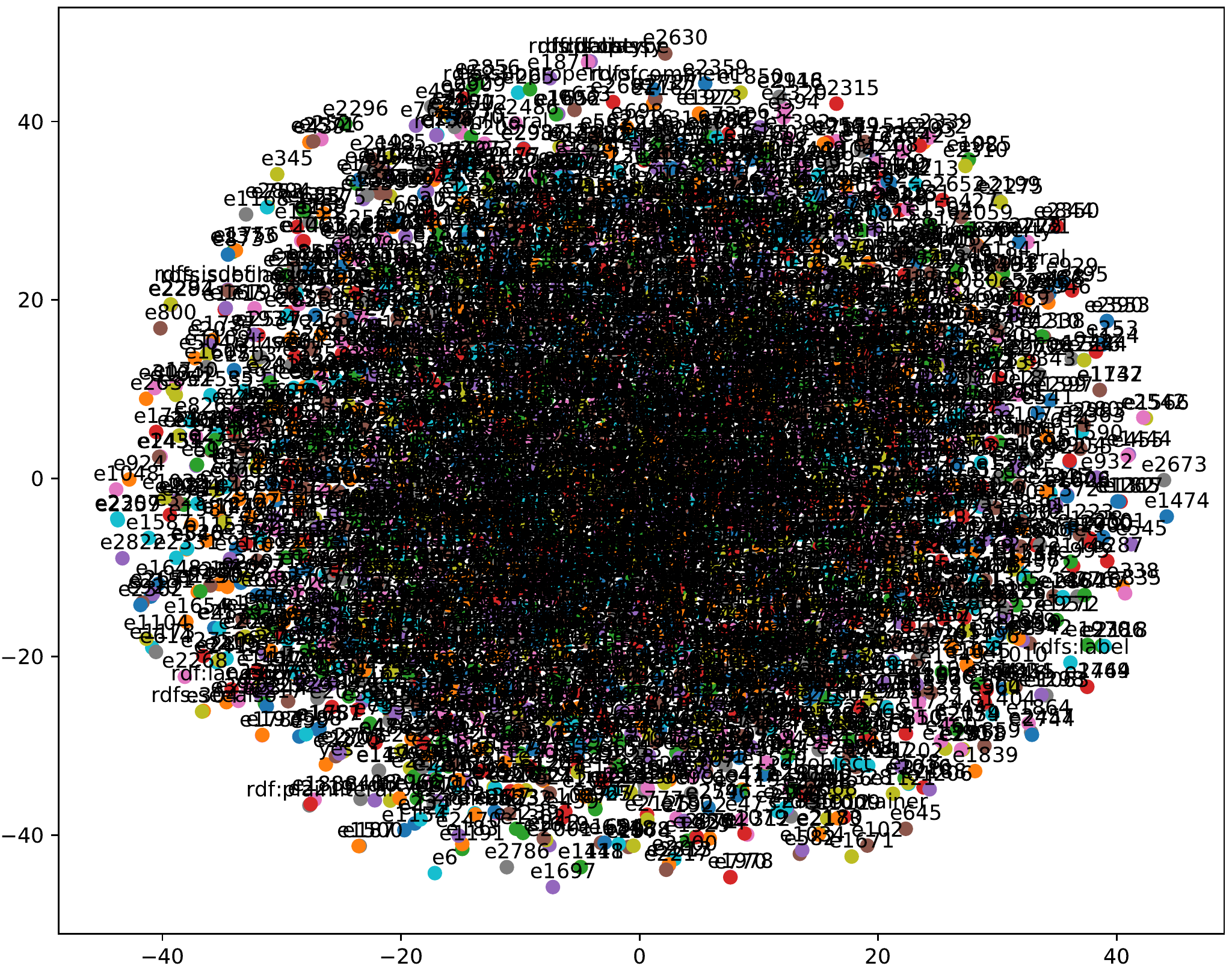}
  \caption{Embeddings for the whole general vocabulary}
  \label{fig:sub2}
\end{subfigure}
\caption{t-SNE projection}
\label{fig:test}
\end{figure}

\begin{figure}[!t]
   \centering
   \includegraphics[width=.5\textwidth]{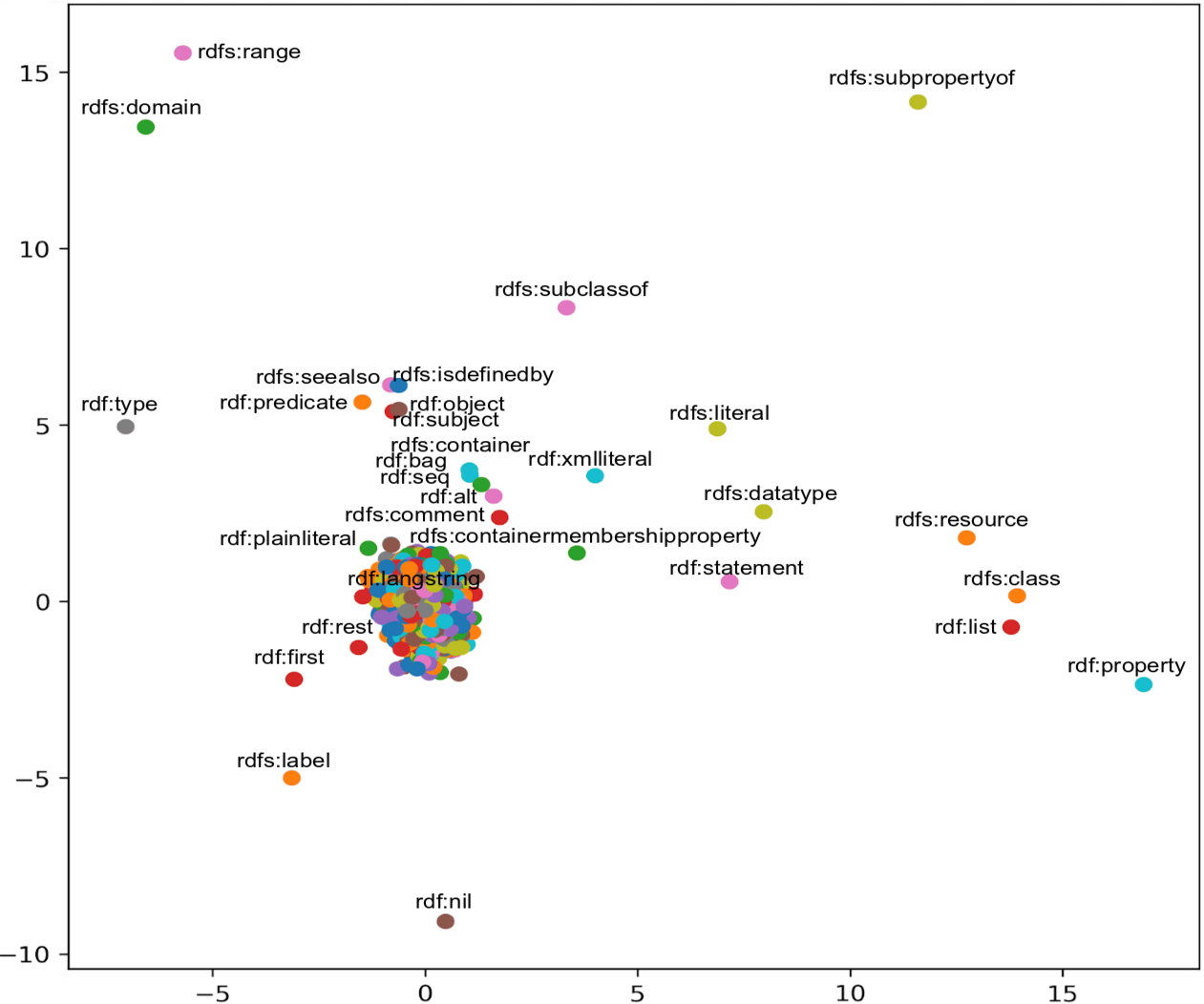}
   \caption{PCA projection of embeddings for the whole general vocabulary}
   \label{}
\end{figure}

\subsection{Ablation Study}
\begin{table*}[]
\begin{adjustbox}{width=1\textwidth}
\begin{threeparttable}
\begin{tabular}{|l|l|l|l|l|l|l|l|l|}
\hline
\multicolumn{1}{|c|}{\multirow{2}{*}{Training Dataset}} & \multirow{2}{*}{Test Dataset}                                                                           & \multicolumn{3}{c|}{Valid Triples Class} & \multicolumn{3}{c|}{Invalid Triples Class} & \multicolumn{1}{|c|}{\multirow{2}{*}{Accuracy}}\\ \cline{3-8} 
\multicolumn{1}{|c|}{}                                  &                                                                                                         & Precision     & Recall    & F-measure    & Precision     & Recall     & F-measure  &   \\ \hline
OWL-Centric Dataset                                  & Linked Data                                                                        & 94            & 97        & 95           & 97            & 93         & 95    &     28   \\
\multicolumn{1}{|c|}{OWL-Centric Dataset (90\%)}        & OWL-Centric Dataset (10\%)                                                                              & 85            & 92        & 88           & 92            & 83         & 87        & 76   \\
OWL-Centric Dataset                                     & OWL-Centric Test Set \tnote{a}                                                                  & 73            & 80        & 75           & 80            & 67         & 71          & 61 \\
OWL-Centric Dataset                                     & Synthetic Data                                                                                          & 52            & 43        & 46           & 51            & 60         & 54        &  51  \\

\hline
\end{tabular}
        \begin{tablenotes}
            \item[a] Completely Different Domain.
        \end{tablenotes}
        \caption{Ablation Study: No Positional Encoding}
\label{my-label}
\end{threeparttable}
\end{adjustbox}
\end{table*}
We perform ablation study where we remove positional encoding from embeddings and compare the results to assess their impact on our results. The idea behind positional encoding is keeping order of elements in the triples into account. Here instead we are using bag of word representations and do not take ordering of elements in each triplet into account. The results of our experiments have been listed in Table 6. As anticipated, removing the positional encoding shows performance decrease for all of our experiments in terms of accuracy. Indeed, through more detailed analysis of the result for our first model, we found that it classifies all our zero-paddings to negative class. That is the explanation for huge gap of the accuracy and f-measure in that model. Nevertheless, surprisingly, but still not hard to appreciate, removing positional encoding would not decrease the performance for some of our experiments substantially. Indeed, this is not practically surprising in light of the fact that orderless representations have always shown tremendous success in natural language processing domain even when order naturally matters.

\subsection{Limitations}
Our work clearly has some limitations. One limitation of our initial approach is that our setting puts a global limit on the size of knowledge graphs a trained system will be able to handle, and required training time can be expected grow super-linearly in the size of the knowledge graphs. We will in fact test the scalability limits of our approach in future, but we are confident that we can reasonably handle knowledge graphs with hundreds of thousands of triples. The contribution of this work, however, is on the fundamental capabilities of our model to perform deductive entailment across knowledge graphs, and thus we have not focused heavily on scalability aspects. Our future work will concentrate on scalability issues more.

Additionally, based on our analysis the reasoning hop length of our real-world datasets are either 2 or 3. This is the case for real-world datasets used in previous studies also. This distribution of training data constrains the capability of our model for learning longer path reasonings. However, due to the high reasoning capacity of memory networks, we are highly confident that the model would be capable of doing that in order of 10s, when it has been trained on complex enough data. Typically, one would expect the number of entailed facts over
the number of inference rule applications to follow a long-tail distribution, which means that in training data, "deep" entailments would be underrepresented, and this may cause a network to not actually acquire deep inference skills. In future works, We will experiment with different synthetic training sets, possibly overrepresenting "deep" entailments, to counter this problem.

Keeping above limitations in mind, our future goal is creating a synthetic dataset with longer reasoning paths for training our model. We would like also to explore the scalability power of our work in future. 

\section{Conclusions}\label{7}
In this paper, we have shown how emulating the symbolic reasoning through sub-symbolic reasoning can lead to a scalable and efficient model capable of transferring its reasoning ability from one domain to another without any re/pre-training or fine-tunning over new domain. To achieve this goal, we have introduced a normalization technique to the representation learning in memory networks. We empirically show our proposed model comfortably beats its unnormalized counterparts. Apart from knowledge graph reasoning, our approach would lend itself well for use by hybrid sub-symbolic symbolic reasoning systems in planning, cognitive systems and robot control. Our study also provides additional considerable insight into not only representation learning for rare or out-of-vocabulary words in general context but also for transfer learning, zero-shot learning and domain adaptation in reasoning domain.

\section{Acknowledgements}
This work is supported by the Ohio Federal Research Network project Human-Centered Big Data.

\bibliographystyle{unsrt}
\bibliography{references}

\begin{thebibliography}{10}

\bibitem{bollacker2008freebase}
Kurt Bollacker, Colin Evans, Praveen Paritosh, Tim Sturge, and Jamie Taylor.
\newblock Freebase: a collaboratively created graph database for structuring
  human knowledge.
\newblock In {\em Proceedings of the 2008 ACM SIGMOD international conference
  on Management of data}, pages 1247--1250. AcM, 2008.

\bibitem{lehmann2015dbpedia}
Jens Lehmann, Robert Isele, Max Jakob, Anja Jentzsch, Dimitris Kontokostas,
  Pablo~N Mendes, Sebastian Hellmann, Mohamed Morsey, Patrick Van~Kleef,
  S{\"o}ren Auer, et~al.
\newblock Dbpedia--a large-scale, multilingual knowledge base extracted from
  wikipedia.
\newblock {\em Semantic Web}, 6(2):167--195, 2015.

\bibitem{dong2014knowledge}
Xin Dong, Evgeniy Gabrilovich, Geremy Heitz, Wilko Horn, Ni~Lao, Kevin Murphy,
  Thomas Strohmann, Shaohua Sun, and Wei Zhang.
\newblock Knowledge vault: A web-scale approach to probabilistic knowledge
  fusion.
\newblock In {\em Proceedings of the 20th ACM SIGKDD international conference
  on Knowledge discovery and data mining}, pages 601--610. ACM, 2014.

\bibitem{min2013distant}
Bonan Min, Ralph Grishman, Li~Wan, Chang Wang, and David Gondek.
\newblock Distant supervision for relation extraction with an incomplete
  knowledge base.
\newblock In {\em Proceedings of the 2013 Conference of the North American
  Chapter of the Association for Computational Linguistics: Human Language
  Technologies}, pages 777--782, 2013.

\bibitem{mccarthy1960programs}
John McCarthy.
\newblock {\em Programs with common sense}.
\newblock RLE and MIT computation center, 1960.

\bibitem{nilsson1991logic}
Nils~J Nilsson.
\newblock Logic and artificial intelligence.
\newblock {\em Artificial intelligence}, 47(1-3):31--56, 1991.

\bibitem{nickel2012factorizing}
Maximilian Nickel, Volker Tresp, and Hans-Peter Kriegel.
\newblock Factorizing yago: scalable machine learning for linked data.
\newblock In {\em Proceedings of the 21st international conference on World
  Wide Web}, pages 271--280. ACM, 2012.

\bibitem{riedel2013relation}
Sebastian Riedel, Limin Yao, Andrew McCallum, and Benjamin~M Marlin.
\newblock Relation extraction with matrix factorization and universal schemas.
\newblock In {\em Proceedings of the 2013 Conference of the North American
  Chapter of the Association for Computational Linguistics: Human Language
  Technologies}, pages 74--84, 2013.

\bibitem{socher2013reasoning}
Richard Socher, Danqi Chen, Christopher~D Manning, and Andrew Ng.
\newblock Reasoning with neural tensor networks for knowledge base completion.
\newblock In {\em Advances in neural information processing systems}, pages
  926--934, 2013.

\bibitem{chang2014typed}
Kai-Wei Chang, Scott Wen-tau Yih, Bishan Yang, and Chris Meek.
\newblock Typed tensor decomposition of knowledge bases for relation
  extraction.
\newblock 2014.

\bibitem{yang2014embedding}
Bishan Yang, Wen-tau Yih, Xiaodong He, Jianfeng Gao, and Li~Deng.
\newblock Embedding entities and relations for learning and inference in
  knowledge bases.
\newblock {\em arXiv preprint arXiv:1412.6575}, 2014.

\bibitem{toutanova2015representing}
Kristina Toutanova, Danqi Chen, Patrick Pantel, Hoifung Poon, Pallavi
  Choudhury, and Michael Gamon.
\newblock Representing text for joint embedding of text and knowledge bases.
\newblock In {\em Proceedings of the 2015 Conference on Empirical Methods in
  Natural Language Processing}, pages 1499--1509, 2015.

\bibitem{trouillon2016complex}
Th{\'e}o Trouillon, Johannes Welbl, Sebastian Riedel, {\'E}ric Gaussier, and
  Guillaume Bouchard.
\newblock Complex embeddings for simple link prediction.
\newblock In {\em International Conference on Machine Learning}, pages
  2071--2080, 2016.

\bibitem{neelakantan2015compositional}
Arvind Neelakantan, Benjamin Roth, and Andrew McCallum.
\newblock Compositional vector space models for knowledge base completion.
\newblock {\em arXiv preprint arXiv:1504.06662}, 2015.

\bibitem{peng2015towards}
Baolin Peng, Zhengdong Lu, Hang Li, and Kam-Fai Wong.
\newblock Towards neural network-based reasoning.
\newblock {\em arXiv preprint arXiv:1508.05508}, 2015.

\bibitem{das2016chains}
Rajarshi Das, Arvind Neelakantan, David Belanger, and Andrew McCallum.
\newblock Chains of reasoning over entities, relations, and text using
  recurrent neural networks.
\newblock {\em arXiv preprint arXiv:1607.01426}, 2016.

\bibitem{weissenborn2016separating}
Dirk Weissenborn.
\newblock Separating answers from queries for neural reading comprehension.
\newblock {\em arXiv preprint arXiv:1607.03316}, 2016.

\bibitem{shen2017reasonet}
Yelong Shen, Po-Sen Huang, Jianfeng Gao, and Weizhu Chen.
\newblock Reasonet: Learning to stop reading in machine comprehension.
\newblock In {\em Proceedings of the 23rd ACM SIGKDD International Conference
  on Knowledge Discovery and Data Mining}, pages 1047--1055. ACM, 2017.

\bibitem{rocktaschel2017end}
Tim Rockt{\"a}schel and Sebastian Riedel.
\newblock End-to-end differentiable proving.
\newblock In {\em Advances in Neural Information Processing Systems}, pages
  3788--3800, 2017.

\bibitem{ling2015finding}
Wang Ling, Tiago Lu{\'\i}s, Lu{\'\i}s Marujo, Ram{\'o}n~Fernandez Astudillo,
  Silvio Amir, Chris Dyer, Alan~W Black, and Isabel Trancoso.
\newblock Finding function in form: Compositional character models for open
  vocabulary word representation.
\newblock {\em arXiv preprint arXiv:1508.02096}, 2015.

\bibitem{bahdanau2017learning}
Dzmitry Bahdanau, Tom Bosc, Stanis{\l}aw Jastrz{\k{e}}bski, Edward
  Grefenstette, Pascal Vincent, and Yoshua Bengio.
\newblock Learning to compute word embeddings on the fly.
\newblock {\em arXiv preprint arXiv:1706.00286}, 2017.

\bibitem{eric2017copy}
Mihail Eric and Christopher~D Manning.
\newblock A copy-augmented sequence-to-sequence architecture gives good
  performance on task-oriented dialogue.
\newblock {\em arXiv preprint arXiv:1701.04024}, 2017.

\bibitem{raghu2018hierarchical}
Dinesh Raghu, Nikhil Gupta, et~al.
\newblock Hierarchical pointer memory network for task oriented dialogue.
\newblock {\em arXiv preprint arXiv:1805.01216}, 2018.

\bibitem{weston2014memory}
Jason Weston, Sumit Chopra, and Antoine Bordes.
\newblock Memory networks. corr abs/1410.3916, 2014.

\bibitem{sukhbaatar2015end}
Sainbayar Sukhbaatar, Jason Weston, Rob Fergus, et~al.
\newblock End-to-end memory networks.
\newblock In {\em Advances in neural information processing systems}, pages
  2440--2448, 2015.

\bibitem{hill2015goldilocks}
Felix Hill, Antoine Bordes, Sumit Chopra, and Jason Weston.
\newblock The goldilocks principle: Reading children's books with explicit
  memory representations.
\newblock {\em arXiv preprint arXiv:1511.02301}, 2015.

\bibitem{bordes2016learning}
Antoine Bordes, Y-Lan Boureau, and Jason Weston.
\newblock Learning end-to-end goal-oriented dialog.
\newblock {\em arXiv preprint arXiv:1605.07683}, 2016.

\bibitem{dodge2015evaluating}
Jesse Dodge, Andreea Gane, Xiang Zhang, Antoine Bordes, Sumit Chopra, Alexander
  Miller, Arthur Szlam, and Jason Weston.
\newblock Evaluating prerequisite qualities for learning end-to-end dialog
  systems.
\newblock {\em arXiv preprint arXiv:1511.06931}, 2015.

\bibitem{mcculloch1943logical}
Warren~S McCulloch and Walter Pitts.
\newblock A logical calculus of the ideas immanent in nervous activity.
\newblock {\em The bulletin of mathematical biophysics}, 5(4):115--133, 1943.

\bibitem{besold2017neural}
Tarek~R Besold, Artur~d'Avila Garcez, Sebastian Bader, Howard Bowman, Pedro
  Domingos, Pascal Hitzler, Kai-Uwe K{\"u}hnberger, Luis~C Lamb, Daniel Lowd,
  Priscila Machado~Vieira Lima, et~al.
\newblock Neural-symbolic learning and reasoning: A survey and interpretation.
\newblock {\em arXiv preprint arXiv:1711.03902}, 2017.

\bibitem{garcez2008neural}
Artur~SD'Avila Garcez, Luis~C Lamb, and Dov~M Gabbay.
\newblock {\em Neural-symbolic cognitive reasoning}.
\newblock Springer Science \& Business Media, 2008.

\bibitem{mccarthy1988epistemological}
John McCarthy.
\newblock Epistemological challenges for connectionism.
\newblock {\em Behavioral and Brain Sciences}, 11(1):44--44, 1988.

\bibitem{towell1994knowledge}
Geoffrey~G Towell and Jude~W Shavlik.
\newblock Knowledge-based artificial neural networks.
\newblock {\em Artificial intelligence}, 70(1-2):119--165, 1994.

\bibitem{hitzler2004logic}
Pascal Hitzler, Steffen H{\"o}lldobler, and Anthony~Karel Seda.
\newblock Logic programs and connectionist networks.
\newblock {\em Journal of Applied Logic}, 2(3):245--272, 2004.

\bibitem{hoelldobler1994massiv}
Steffen Hoelldobler and Yvonne Kalinke.
\newblock Ein massiv paralleles modell f{\"u}r die logikprogrammierung.
\newblock In {\em WLP}, pages 89--92, 1994.

\bibitem{shastri1999advances}
Lokendra Shastri.
\newblock Advances in shruti—a neurally motivated model of relational
  knowledge representation and rapid inference using temporal synchrony.
\newblock {\em Applied Intelligence}, 11(1):79--108, 1999.

\bibitem{shastri2007shruti}
Lokendra Shastri.
\newblock Shruti: A neurally motivated architecture for rapid, scalable
  inference.
\newblock In {\em Perspectives of Neural-Symbolic Integration}, pages 183--203.
  Springer, 2007.

\bibitem{gust2007learning}
Helmar Gust, Kai-Uwe K{\"u}hnberger, and Peter Geibel.
\newblock Learning models of predicate logical theories with neural networks
  based on topos theory.
\newblock In {\em Perspectives of Neural-Symbolic Integration}, pages 233--264.
  Springer, 2007.

\bibitem{bader2008connectionist}
Sebastian Bader, Pascal Hitzler, and Steffen H{\"o}lldobler.
\newblock Connectionist model generation: A first-order approach.
\newblock {\em Neurocomputing}, 71(13-15):2420--2432, 2008.

\bibitem{bader2007fully}
Sebastian Bader, Pascal Hitzler, Steffen H{\"o}lldobler, Andreas Witzel, et~al.
\newblock A fully connectionist model generator for covered first-order logic
  programs.
\newblock In {\em IJCAI}, pages 666--671, 2007.

\bibitem{asai2017classical}
Masataro Asai and Alex Fukunaga.
\newblock Classical planning in deep latent space: Bridging the
  subsymbolic-symbolic boundary.
\newblock {\em arXiv preprint arXiv:1705.00154}, 2017.

\bibitem{donadello2017logic}
Ivan Donadello, Luciano Serafini, and Artur~d'Avila Garcez.
\newblock Logic tensor networks for semantic image interpretation.
\newblock {\em arXiv preprint arXiv:1705.08968}, 2017.

\bibitem{hohenecker2018ontology}
Patrick Hohenecker and Thomas Lukasiewicz.
\newblock Ontology reasoning with deep neural networks.
\newblock {\em arXiv preprint arXiv:1808.07980}, 2018.

\bibitem{maknideep}
Bassem Makni and James Hendler.
\newblock Deep learning for noise-tolerant rdfs reasoning.

\bibitem{serafini2016learning}
Luciano Serafini and Artur S~d’Avila Garcez.
\newblock Learning and reasoning with logic tensor networks.
\newblock In {\em Conference of the Italian Association for Artificial
  Intelligence}, pages 334--348. Springer, 2016.

\bibitem{serafini2016logic}
Luciano Serafini and Artur~d'Avila Garcez.
\newblock Logic tensor networks: Deep learning and logical reasoning from data
  and knowledge.
\newblock {\em arXiv preprint arXiv:1606.04422}, 2016.

\bibitem{nguyenconvolutional}
Dai~Quoc Nguyen, Dat~Quoc Nguyen, Tu~Dinh Nguyen, and Dinh Phung.
\newblock A convolutional neural network-based model for knowledge base
  completion and its application to search personalization.
\newblock {\em Semantic Web}, (Preprint):1--14.

\bibitem{grefenstette2015learning}
Edward Grefenstette, Karl~Moritz Hermann, Mustafa Suleyman, and Phil Blunsom.
\newblock Learning to transduce with unbounded memory.
\newblock In {\em Advances in Neural Information Processing Systems}, pages
  1828--1836, 2015.

\bibitem{world2014rdf}
World Wide~Web Consortium et~al.
\newblock Rdf 1.1 concepts and abstract syntax.
\newblock 2014.

\bibitem{hitzler2009foundations}
Pascal Hitzler, Markus Krotzsch, and Sebastian Rudolph.
\newblock {\em Foundations of semantic web technologies}.
\newblock Chapman and Hall/CRC, 2009.

\bibitem{hitzler2009owl}
Pascal Hitzler, Markus Kr{\"o}tzsch, Bijan Parsia, Peter~F Patel-Schneider, and
  Sebastian Rudolph.
\newblock Owl 2 web ontology language primer.
\newblock {\em W3C recommendation}, 27(1):123, 2009.

\bibitem{guu2015traversing}
Kelvin Guu, John Miller, and Percy Liang.
\newblock Traversing knowledge graphs in vector space.
\newblock {\em arXiv preprint arXiv:1506.01094}, 2015.

\bibitem{xiong2017deeppath}
Wenhan Xiong, Thien Hoang, and William~Yang Wang.
\newblock Deeppath: A reinforcement learning method for knowledge graph
  reasoning.
\newblock {\em arXiv preprint arXiv:1707.06690}, 2017.

\bibitem{rocktaschel2015injecting}
Tim Rockt{\"a}schel, Sameer Singh, and Sebastian Riedel.
\newblock Injecting logical background knowledge into embeddings for relation
  extraction.
\newblock In {\em Proceedings of the 2015 Conference of the North American
  Chapter of the Association for Computational Linguistics: Human Language
  Technologies}, pages 1119--1129, 2015.

\bibitem{bordes2011learning}
Antoine Bordes, Jason Weston, Ronan Collobert, Yoshua Bengio, et~al.
\newblock Learning structured embeddings of knowledge bases.
\newblock In {\em AAAI}, volume~6, page~6, 2011.

\bibitem{bordes2013translating}
Antoine Bordes, Nicolas Usunier, Alberto Garcia-Duran, Jason Weston, and Oksana
  Yakhnenko.
\newblock Translating embeddings for modeling multi-relational data.
\newblock In {\em Advances in neural information processing systems}, pages
  2787--2795, 2013.

\bibitem{lin2015learning}
Yankai Lin, Zhiyuan Liu, Maosong Sun, Yang Liu, and Xuan Zhu.
\newblock Learning entity and relation embeddings for knowledge graph
  completion.
\newblock In {\em AAAI}, volume~15, pages 2181--2187, 2015.

\bibitem{wang2014knowledge}
Zhen Wang, Jianwen Zhang, Jianlin Feng, and Zheng Chen.
\newblock Knowledge graph embedding by translating on hyperplanes.
\newblock In {\em AAAI}, volume~14, pages 1112--1119, 2014.

\bibitem{cai2018comprehensive}
Hongyun Cai, Vincent~W Zheng, and Kevin Chang.
\newblock A comprehensive survey of graph embedding: problems, techniques and
  applications.
\newblock {\em IEEE Transactions on Knowledge and Data Engineering}, 2018.

\bibitem{madotto2018mem2seq}
Andrea Madotto, Chien-Sheng Wu, and Pascale Fung.
\newblock Mem2seq: Effectively incorporating knowledge bases into end-to-end
  task-oriented dialog systems.
\newblock {\em arXiv preprint arXiv:1804.08217}, 2018.

\bibitem{cheatham2018geolink}
Michelle Cheatham, Adila Krisnadhi, Reihaneh Amini, Pascal Hitzler, Krzysztof
  Janowicz, Adam Shepherd, Tom Narock, Matt Jones, and Peng Ji.
\newblock The geolink knowledge graph.
\newblock {\em Big Earth Data}, pages 1--13, 2018.

\bibitem{janowicz2014five}
Krzysztof Janowicz, Pascal Hitzler, Benjamin Adams, Dave Kolas, II~Vardeman,
  et~al.
\newblock Five stars of linked data vocabulary use.
\newblock {\em Semantic Web}, 5(3):173--176, 2014.

\bibitem{sam2018quality}
Stella Sam, Pascal Hitzler, and Krzysztof Janowicz.
\newblock On the quality of vocabularies for linked dataset papers published in
  the semantic web journal, 2018.

\bibitem{das2017go}
Rajarshi Das, Shehzaad Dhuliawala, Manzil Zaheer, Luke Vilnis, Ishan Durugkar,
  Akshay Krishnamurthy, Alex Smola, and Andrew McCallum.
\newblock Go for a walk and arrive at the answer: Reasoning over paths in
  knowledge bases using reinforcement learning.
\newblock {\em arXiv preprint arXiv:1711.05851}, 2017.

\bibitem{yang2017differentiable}
Fan Yang, Zhilin Yang, and William~W Cohen.
\newblock Differentiable learning of logical rules for knowledge base
  reasoning.
\newblock In {\em Advances in Neural Information Processing Systems}, pages
  2319--2328, 2017.

\bibitem{maaten2008visualizing}
Laurens van~der Maaten and Geoffrey Hinton.
\newblock Visualizing data using t-sne.
\newblock {\em Journal of machine learning research}, 9(Nov):2579--2605, 2008.

\end{thebibliography}

\end{document}